# Towards Explainable Indoor Localization: Interpreting Neural Network Learning on Wi-Fi Fingerprints Using Logic Gates

Danish Gufran and Sudeep Pasricha
Department of Electrical and Computer Engineering
Colorado State University, Fort Collins, CO, USA
{danish.gufran, sudeep}@colostate.edu

*Abstract*— **Indoor localization using deep learning (DL) has demonstrated strong accuracy in mapping Wi-Fi RSS fingerprints to physical locations; however, most existing DL frameworks function as black-box models, offering limited insight into how predictions are made or how models respond to real-world noise over time. This lack of interpretability hampers our ability to understand the impact of temporal variations—caused by environmental dynamics—and to adapt models for long-term reliability. To address this, we introduce LogNet, a novel logic gate-based framework designed to interpret and enhance DL-based indoor localization. LogNet enables transparent reasoning by identifying which access points (APs) are most influential for each reference point (RP) and reveals how environmental noise disrupts DL-driven localization decisions. This interpretability allows us to trace and diagnose model failures and adapt DL systems for more stable long-term deployments. Evaluations across multiple real-world building floorplans and over two years of temporal variation show that LogNet not only interprets the internal behavior of DL models but also improves performance—achieving up to 1.1× to 2.8× lower localization error, 3.4× to 43.3× smaller model size, and 1.5× to 3.6× lower latency compared to prior DL-based models.**

*Keywords—Explainable AI, Indoor Localization, Temporal Variations, Logic Gate Networks.*

## I. Introduction

Indoor localization has become a cornerstone of modern context-aware technologies, enabling applications in robotics, augmented and virtual reality (AR/VR), asset tracking, and emergency response. One of the earliest indoor localization system, "*The Active Badge Location System*" introduced in 1992 [1], relied on infrared (IR) pulses emitted by wearable badges and captured by stationary IR receivers [1]. However, its reliance on line-of-sight (LoS) for IR communications rendered it ineffective in large-scale environments. As indoor environments have grown more dynamic and complex, researchers have increasingly turned to artificial intelligence (AI) to enhance localization robustness and scalability [2]. This paradigm shift has driven widespread adoption across industries, with companies such as Apple, Google, Microsoft, HPE, Quuppa, and Zebra investing heavily in AI-powered indoor localization systems [3]. Reflecting this momentum, the global indoor localization market was valued at $11.9 billion in 2024, with strong forecasts for continued growth driven by the demand for high-precision and intelligent indoor localization frameworks [4].

Among AI techniques, neural network based Deep Learning (DL) models have shown promise in capturing spatial patterns within Radio Frequency (RF) data for indoor localization [5]. While indoor localization systems may use various RF technologies—such as Bluetooth, Ultra-Wideband (UWB), or cellular signals—Wi-Fi remains the most widely adopted due to its pervasive presence and seamless integration with mobile devices. Wi-Fi-based localization relies on constructing high-dimensional "fingerprints" from Received Signal Strength (RSS) values measured across multiple Access Points (APs) per Reference Point (RP) location [6]. DL models have been favored for their ability to learn non-linear decision boundaries across high-dimensional RSS fingerprints. However, they typically function as opaque black-box systems, offering limited interpretability into how RSS from specific APs can influence localization predictions [7].

This black-box assumption becomes problematic in dynamic indoor environments, where noise introduced by factors such as human movement, signal shadowing, and multipath signal propagation can degrade model accuracy [8]. While some of this noise may have short-term effects, long-term environmental changes such as permanent furniture rearrangement and aging/replacement of Wi-Fi transceivers lead to what is known as *temporal variations* that distort the RSS fingerprint distributions over time [2], [8]. In DL, these variations can cause learned decision boundaries between RPs to break down, leading to higher localization errors and unreliable long-term performance [9]. More critically, the lack of interpretability in DL models leaves us unable to identify *Which APs are most influential for a given RP?* and *How does the model form decision boundaries between RPs?* Without answers to these questions, diagnosing failures and improving model robustness becomes exceptionally challenging.

To address these important issues, we introduce LogNet, a novel logic gate-based network that not only interprets the internal representations of DL models but also enhances their robustness to temporal variations. LogNet leverages logic gate operations to reveal the most influential APs per RP and exposes how environmental changes disrupt classification. These insights enable us to enhance DL models and extend their usability over time. Our work aims to answer three key research questions: **RQ1** – *What underlying factors cause DL-based indoor localization models to deteriorate under temporal variation?* **RQ2** – *How can identifying the most influential APs for each RP help in enhancing model robustness against temporal variations?* **RQ3** – *In what ways can logic gate networks be employed to both interpret DL decision processes and improve their resilience to temporal variations?* Our novel contributions are as follows:

- We introduce LogNet, a framework that brings interpretability to DL-based indoor localization.
- We demonstrate how LogNet can be used to identify failures in DL models due to temporal variation and improve their resilience.
- We perform extensive evaluations on real-world datasets collected over two years across buildings, showing that LogNet outperforms existing DL methods in both interpretability and long-term localization performance.

## II. RELATED WORKS

Indoor localization techniques have progressed rapidly over the past decade, evolving from traditional geometric methods to DL-based approaches, driven by the need for greater accuracy and robustness in real-world environments. Classical systems such as TDoA-Loc [10], AngLoc [11], RSSI-Trilat [12], and BGI-Trilateration [13] leveraged geometric principles—like time-difference-of-arrival (TDoA) and angle-of-arrival (AoA)—to estimate a user's position relative to known RPs. These models offered spatially interpretable decision boundaries, making them attractive for explainable deployments. However, their practical utility was constrained by the need for clear LoS, susceptibility to multipath propagation, and signal shadowing—common challenges in indoor environments.

To address these limitations, fingerprinting-based methods emerged, including RADAR [14]. These approaches construct a database of RSS fingerprints collected across RPs and use similarity metrics such as k-nearest neighbors (k-NN) or Euclidean distance (ED) to estimate positions. Compared to geometric methods, fingerprinting techniques demonstrated better adaptability in non-LoS conditions. However, fingerprinting techniques lack the interpretability of geometric models. Since prediction is based on similarity matching, it is difficult to isolate the contribution of an individual AP to localization decisions [9]. Furthermore, these methods assume noise affects all APs uniformly—i.e., a Euclidean view of the fingerprint vector. In contrast, real-world noise is often non-uniform across APs, leading to a more complex, non-Euclidean distribution (further discussed in Section III).

To overcome these shortcomings, recent advancements have turned to DL algorithms, particularly deep neural networks (DNNs), which are adept at learning complex nonlinear mappings from high-dimensional RSS fingerprints to RP locations. Several models—such as DNNLOC [15], ISDL [16], and HWF [17]—have demonstrated improved accuracy over similarity-based fingerprinting methods. These DNNs typically adopt one of three core architectural patterns [18]: 1) DNN-Dense, where all hidden layers maintain uniform dimensionality; 2) DNN-UpSample, which progressively increases dimensionality in deeper layers; and 3) DNN-DownSample, where hidden layers compress the feature space successively (details in Section IV).

Despite these advancements, all DNN-based architectures share a key limitation: performance deterioration over long-term temporal variations. These variations which occur over time (weeks to months after initial deployment) disrupt the learned latent space representations and decision boundaries between RP classes. Compounding this issue is the black-box nature of DNNs, which hinders interpretability and makes it difficult to identify which APs are most influential for any given RP [19]. Moreover, DNNs often implicitly process fingerprint vectors under the assumption of uniform (Euclidean) noise, further exacerbating their failure under real-world, non-Euclidean conditions.

Addressing this challenge requires a deeper understanding of how DNNs encode RSS data in their latent space (decision boundary space between classes) and how these encodings evolve with time. This insight is essential not only for diagnosing model failures but also for improving resilience against temporal variation. This is the main motivation behind LogNet. LogNet uses logic gates to explicitly trace the contribution of individual APs to model predictions and identify how decision boundaries are violated due to temporal variations (discussed in Section III). In doing so, it provides actionable insights into existing DNN limitations for indoor localization and establishes a foundation for future architectures that balance performance with interpretability in the presence of real-world non-Euclidean noise.

## III. INTERPRETING DL FOR INDOOR LOCALIZATION

In this section, we first analyze the structure of noise distribution in Wi-Fi RSS fingerprint vectors caused by temporal variations and demonstrate how conventional assumptions of uniformity fail in real-world settings. We then introduce logic gates as interpretable computational primitives and explore how their behavior can be harnessed to understand and model the internal learning processes of DNNs for indoor localization.

### A. Structural Analysis of Wi-Fi RSS Fingerprints

As a first step, we analyze the structural characteristics of Wi-Fi RSS fingerprints ($F$) and the temporal noise patterns that arise over time due to environmental dynamics. Traditionally, $F$ is assumed to follow a Euclidean (ED) structure, where noise is distributed uniformly across all APs. However, real-world deployments often exhibit non-uniform noise patterns, leading to a non-Euclidean (non-ED) structure. As illustrated in Figure 1 and formalized in Equation (1), in an ED structure, a constant noise term ($\delta$) is assumed and applied across all APs. In contrast, Equation (2) describes the non-ED structure, where each AP is subject to its own unique noise term, capturing the uneven impact of temporal variations across APs.

$$F_{ED} = [(\delta, RSS_{AP:0}), (\delta, RSS_{AP:1}), \dots, (\delta, RSS_{AP:N})] \quad (1)$$

$$F_{Non-ED} = [(\delta_0, RSS_{AP:0}), (\delta_1, RSS_{AP:1}), \dots, (\delta_N, RSS_{AP:N}) \quad (2)$$

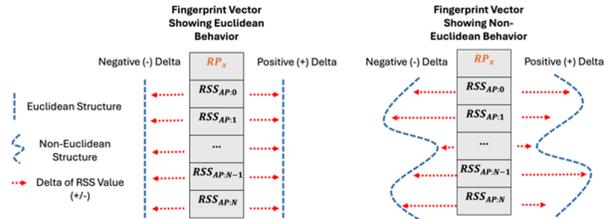

Figure 1: Euclidean vs. Non-Euclidean fingerprint structures.

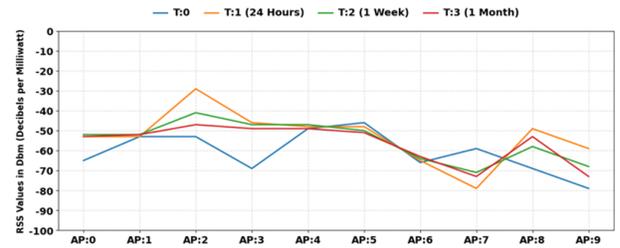

Figure 2: Temporal variation of RSS values across the first 10 APs.

To empirically verify this non-ED behavior, we collected RSS fingerprints from the same smartphone device and RP over four time points: T:0 (initial), T:1 (after 24 hours), T:2 (after 1 week), and T:3 (after 1 month). Figure 2 shows the temporal fluctuations in RSS across the first 10 APs within $F$. While APs like AP:1, AP:4, AP:5, and AP:6 remain relatively stable, others display significant variation, confirming the presence of non-ED noise. This structural insight challenges the ED assumptions inherent in most DNN-based localization models.

TABLE 1: TRUTH TABLE OF LOGIC GATES

| AND Gate | | | OR Gate | | | NAND Gate | | |
|---|---|---|---|---|---|---|---|---|
| X | Y | Z | X | Y | Z | X | Y | Z |
| 0 | 0 | 0 | 0 | 0 | 0 | 0 | 0 | 1 |
| 0 | 1 | 0 | 0 | 1 | 1 | 0 | 1 | 1 |
| 1 | 0 | 0 | 1 | 0 | 1 | 1 | 0 | 1 |
| 1 | 1 | 1 | 1 | 1 | 1 | 1 | 1 | 0 |

| NOR Gate | | | XOR Gate | | | XNOR Gate | | |
|---|---|---|---|---|---|---|---|---|
| X | Y | Z | X | Y | Z | X | Y | Z |
| 0 | 0 | 1 | 0 | 0 | 0 | 0 | 0 | 1 |
| 0 | 1 | 0 | 0 | 1 | 1 | 0 | 1 | 0 |
| 1 | 0 | 0 | 1 | 0 | 1 | 1 | 0 | 0 |
| 1 | 1 | 0 | 1 | 1 | 0 | 1 | 1 | 1 |

*B. Logic Gates for Indoor Localization*

To enable interpretable reasoning in DL-based indoor localization, we introduce logic gates as foundational components for decision-making. Each gate receives two inputs derived from the RSS fingerprint vector—representing whether two APs are active (1) or inactive (0) based on a defined threshold (discussed in Section IV). These gates evaluate binary relationships between pairs of APs and form the basis of simple and transparent decision rules that can be visualized and interpreted in real time. The truth table for each logic gate is shown in Table 1. The behavior of various types of gates that we considered are summarized below:

- **AND Gate:** Outputs 1 only when both adjacent APs are active, distinguishing regions in the fingerprint, where both APs exhibit strong RSS patterns (co-occurrence) from those where they do not.

$$Z = X.Y \quad (3)$$

- **OR Gate:** Outputs 1 if at least one of the adjacent APs is active, distinguishing regions in the fingerprint, where there is some AP visibility and separates them from dead zone patterns (both APs showing weak RSS).

$$Z = X + Y \quad (4)$$

- **NAND Gate:** Outputs 1 unless both adjacent APs are active, distinguishing regions in the fingerprint, where at least one AP is weak and separate zones with co-occurrence patterns.

$$Z = 1 - (X.Y) \quad (5)$$

- **NOR Gate:** Outputs 1 when both adjacent APs are inactive, distinguishing regions in the fingerprint showing dead zones patterns from those where they do not.

$$Z = 1 - (X + Y) \quad (6)$$

- **XOR Gate:** Outputs 1 when exactly one of the two APs is active, distinguishing regions in the fingerprint, where only one AP is strong and separates regions showing co-activation and dead zone patterns.

$$Z = X.(1-Y) + (1-X).Y \quad (7)$$

- **XNOR Gate:** Outputs 1 when both APs are in the same state—either both active or both inactive, distinguishing regions in the fingerprint, where the APs either show co-activation or dead zone patterns.

$$Z = X.Y + (1-X).(1-Y) \quad (8)$$

By mapping binarized RSS relationships through these gates, we uncover interpretable logic patterns that contribute to a location decision. These gates form the building blocks of the proposed LogNet architecture, allowing us to reason about how RSS patterns are internally processed and how decision boundaries emerge, as discussed next.

## IV. THE LOGNET FRAMEWORK

The LogNet framework introduces an interpretable alternative to conventional DNN architectures for indoor localization by constructing a multi-layer network composed entirely of binary logic gates. As discussed in Section III, each gate operates on a pair of binarized RSS values—designated as either active (1) or inactive (0)—and performs a predefined logical operation (e.g., AND, OR, NAND). This architecture enables LogNet to offer rule-based reasoning while learning spatial relationships among APs in a transparent and diagnosable (traceable) manner.

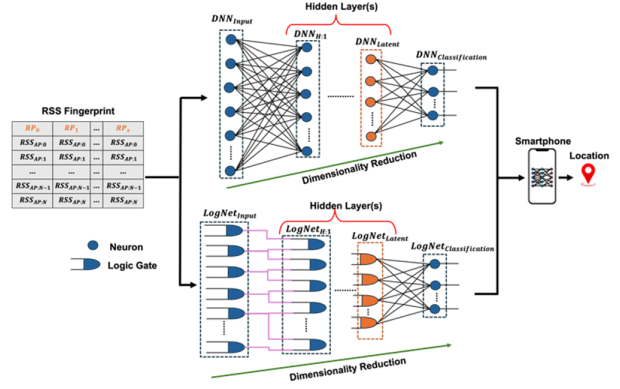

Figure 3: The LogNet and DNN-DownSample architectures.

To design LogNet, we first adopt a baseline DNN-based architecture commonly used in indoor localization, which we term **DNN-DownSample**. This architecture consists of an input layer ($DNN_{Input}$) for RSS fingerprints, followed by a sequence of *hidden* layers ($DNN_{H:N}$), where $N$ is the number of layers with progressively decreasing neurons—thereby performing dimensionality reduction—and a final *Softmax* output layer for RP classification, as shown in figure 3. This DNN-DownSample architecture has been widely used in prior DNN-based indoor localization works, including DNNLOC [15], SANGRIA [20], and AUTLOC [21] due to its effectiveness in compressing high-dimensional RSS vectors and mitigating the effects of short-term temporal variations. Each *hidden* layer reduces the feature space, often by halving the neuron count from the previous layer, resulting in a compact latent space representation. As shown in Figure 3, each neuron in this architecture computes a weighted sum over its inputs followed by a non-linear activation function:

$$Z = \sigma(\sum_i w_i . x_i + B_i) \quad (9)$$

where $x_i$ are the input features, $w_i$ the trained weights, $B_i$ the bias, and $\sigma$ the activation function (ReLu). This model is trained end-to-end using backpropagation to minimize the classification loss (*Sparse Categorical Crossentropy*). However, this process results in an opaque latent space, making it difficult to trace how specific APs influence the final indoor localization decisions (shown in Section V.C).

LogNet addresses this limitation by replacing these opaque transformations (neurons) with logic gates that explicitly model pairwise AP interactions. As shown in

Figure 3, the process begins by binarizing the RSS fingerprint vectors ($X_{train}$). All RSS values are first normalized to the [0, 1] range and then a threshold is applied using a fixed value $\emptyset$ (e.g., 0.5). Any normalized RSS above $\emptyset$ is treated as active (1), and values below are considered inactive (0). This yields a binary fingerprint vector ($B_{train}$) for each training sample. In the first logic layer ($L_{input}$), the binary RSS values are grouped into adjacent pairs. Each pair is passed through a selected logic gate ($G$), producing a single binary output ($G_{output}$). As a result, the output dimension of the first logic layer is half the size of the input (zero-padding applied to maintain even length). The outputs of $L_{input}$ become the inputs to the next layer ($L_k$), where the same pairwise gate operation is repeated. This process continues for a predefined number of layers (*hidden*), forming a DNN-DownSample-like hierarchy where each layer compresses the feature space while retaining key activation patterns. The final output after *hidden* defines the LogNet latent space ($L_{latent}$) —a compact binary encoding that captures discriminative information across APs. $L_{latent}$ is then passed to a fully connected *Softmax* layer, which maps it to the final predicted RP class. *To the best of our knowledge, this is the first work to employ logic gates for interpreting AI models in indoor localization.*

---
**ALGORITHM 1:** The LogNet Algorithm
---
**Input**: RSS fingerprints $X_{train}$, RP classes $Y_{train}$, threshold $\theta$, gate type $G$, number of logic layers *hidden*
**Output**: LogNet latent space $L_{latent}$, Trained LogNet Model $M$
1: $B_{train} \leftarrow Binarize(X_{train}, \emptyset)$ :$B[i]$=1 **if** $X_{train}[i] \geq \emptyset$, **else** 0
2: $L_{input} \leftarrow B_{train}$
3: **for** $k$ = 1 to *hidden* **do**
4:    **for** sample $SMP \in$ layer $L_k$ **do**
5:       **if** len($SMP$) mod 2 $\neq$ 0
6:          $SMP \leftarrow$ append($SMP$, 0)
7:       $Temp \leftarrow [\,]$
8:       **for** $i$ = 0 to len($SMP$) $-$ 1 **do**
9:          $G_{output} \leftarrow ApplyGate(SMP[i], SMP[i+1], G)$
10:         $Temp$.append($G_{output}$)
11:       $L_{k+1}$.append($Temp$)
12:    $L_{saved} \leftarrow L_{k+1}$
13: $L_{latent} \leftarrow L_{saved}$
14: $M \leftarrow$ Train *Softmax* layer ($L_{latent}, Y_{train}$)
---

The complete training process—binarization, logic layer construction, and classifier training—is formalized in Algorithm 1. First, RSS fingerprints $X_{train}$ are binarized using threshold $\emptyset$ to produce binarized fingerprints $B_{train}$, which are then assigned as input to $L_{input}$ (lines 1–2). Next, the logic gate network is constructed for all *hidden* layers (lines 3-12). In each layer $L_k$, sample vectors *(SMP)* are formed by grouping 2 adjacent pairs of inputs: directly from $B_{train}$ when $k$ = 1, or from the outputs of the previous layer when $k$ > 1 (line 4). Zero-padding is applied to ensure even pairing (lines 5–6). A list $Temp$ is initialized to store $G_{output}$ (line 7), and binary gate operations are applied pairwise (lines 8–10) using the selected gate $G$. The resulting encoded vector ($Temp$) is added to the next layer $L_{k+1}$ (line 11), which becomes the input to the subsequent logic layer $L_{saved}$ (line 12). After all *hidden* layers are applied, the final output $L_{latent}$ is defined (line 13). Finally, a *Softmax* classifier is trained on $L_{latent}$ with labels $Y_{train}$ (line 14), using backpropagation similar to training DNNs for classification.

## V. EXPERIMENTS

### A. Experimental Setup

In this section, we outline the experimental setup used to evaluate the LogNet framework. The objectives are: 1) to demonstrate how DNN-based indoor localization models can be interpreted; 2) to empirically address the research questions: **RQ1**, **RQ2**, and **RQ3** from Section I, and 3) to assess the performance improvements that LogNet provides over standard DNNs.

We collected real-world Wi-Fi RSS fingerprints using six smartphones from different manufacturers. The devices are: BLU Vivo 8, HTC U11, Samsung Galaxy S7, LG V20, Motorola Z2, and Oneplus 3. Data was gathered across two buildings—Building 1 and Building 2—with distinct architectural and material characteristics. Building 1 consisted of wood and cement interiors and contained 61 RPs with 164 APs per RP, while Building 2 featured a mix of metal and wooden elements and contained 48 RPs with 218 APs per RP. To ensure realistic conditions, fingerprints were captured during regular work hours without restricting human presence. Each collection instance (CI) included six fingerprints per RP, spanning 10 CIs collected over 2 years. The first three CIs (CI:0-2) were collected on the same day at different times—morning (8 A.M.), afternoon (3 P.M.), and night (9 P.M.)—to account for short term temporal variations. Subsequent CIs were taken at increasing temporal intervals: after 24 hours, 1 week, 1 month, 3 months, 6 months, 1 year, and 2 years (CI:3–9). Data was split using a 5:1 ratio—five fingerprints per RP for training and one for online testing—ensuring consistency in evaluation. A 1-meter RP spacing was maintained across the path. *Additionally, we have open-sourced our data [22] to support reproducibility and research in indoor localization.*

With this dataset [22], we proceed to evaluate both DNN-DownSample and LogNet under identical training settings. Both models used a learning rate of 0.01 using the Adam optimizer. DNN-DownSample was trained for 500 epochs due to its fully trainable architecture, while LogNet needed only 150, since its rule-based logic gates generate the latent space without training. Only the final classification layer is trained. In our subsequent experiments, we focus on the DNN-DownSample model as the baseline for comparison since it structurally aligns with LogNet's layered reduction pattern, we also include a study comparing LogNet with prior DNN-based architectures used in indoor localization (Section V.D). DNN-DownSample, like other DNN variants offers limited interpretability, making it difficult to trace how specific APs contribute to decisions. In contrast, LogNet exposes AP-level influence through logic gate activations, enabling us to identify the discriminative features (APs) per RP. This insight not only allows model transparency but also enhances localization robustness under temporal variations.

### B. Robustness to Temporal Variations: LogNet vs. DNN

We evaluate the performance of six LogNet variants—each using a fixed logic gate (AND, OR, NAND, NOR, XOR, XNOR)—against the baseline DNN-DownSample architecture, with all models containing 1 hidden layer. All models are trained on the training five fingerprints per RP from CI:0 and tested on the holdout sixth fingerprint (single fingerprint per RP) from CI:0 (to compare performance without any temporal variations). As shown in Figure 4, all LogNet variants significantly outperform the DNN-DownSample baseline. In particular, LogNet-NOR achieves the best performance (lowest localization errors) with a mean localization error of 2.75 meters, followed by LogNet-NAND at 2.89 meters. In contrast, DNN-DownSample records a higher error of 5.81 meters. Compared to DNN-DownSample, LogNet-AND, OR, NAND, NOR, XOR, and XNOR achieve 1.39×, 1.48×, 1.50×, 1.52×, 1.08×, and 1.10× lower errors, respectively. These performance gains can be attributed to the logic gate-based architecture of LogNet, where deterministic decision rules filter noise and capture meaningful activation relationships among APs (as discussed in Section III.B).

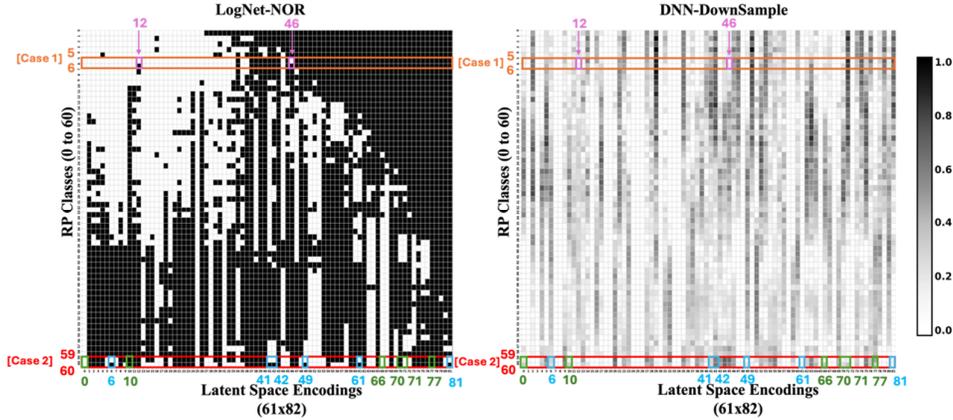

Figure 6: Latent space encodings of LogNet-NOR and DNN-DownSample on Building 1 across all RPs at CI:0.

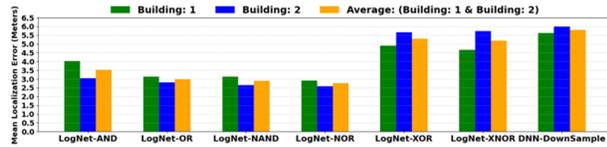

Figure 4: Mean localization error across six LogNet variants and DNN-DownSample at CI:0 (without temporal variations).

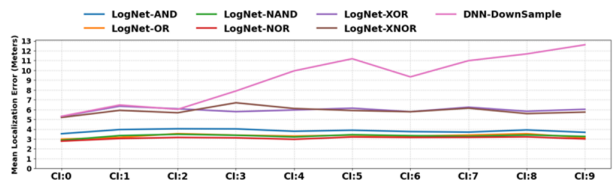

Figure 5: Mean localization error across six LogNet variants and DNN-DownSample over temporal variations (CI:0-9).

To assess temporal robustness (both short and long-term), Figure 5 evaluates all models when trained at CI:0 and tested across CIs:0-9. From CI:0 to CI:2 (collected on the same day), DNN-DownSample performs comparably to LogNet-XOR, indicating that DNNs possess short-term temporal resilience. However, beyond CI:2, DNN error rises steadily peaking at CI:9, showing performance deterioration over long-term temporal variations. This deterioration stems from the DNN's implicit assumption of ED noise distributions, which fails under non-ED distributions. This distorts the DNN's decision boundaries in the latent space, leading to unstable predictions. However, all LogNet variants, particularly LogNet-NOR, maintains a flatter error profile with up to 1.76× lower errors than DNN-DownSample. These results directly answer **RQ1**, showing that temporal variation—especially in the form of non-ED noise—corrupts the latent space representations of DNNs, degrading their long-term accuracy. LogNet's rule-based logic enables it to maintain reliable performance even under long term variations.

### C. Latent Space Interpretability and Failure Traceability

Next, we compare the latent space representations of the best-performing logic gate variant from Section V.B—LogNet-NOR—against the DNN-DownSample model. Both models use a single hidden layer, resulting in latent space encodings with half the dimensionality of the input RSS fingerprint. For this analysis, we focus on Building 1, which contains 61 RPs and 164 APs per RP, resulting in input vectors of shape (61×164) and latent space encoding of shape (61×82). Figure 6 visualizes these encodings across all RP classes, where each row corresponds to an RP and each column to its corresponding latent space encodings. As shown in Figure 6, DNN-DownSample produces continuous-valued latent encodings (ranging from 0 to 1), while LogNet generates strict binary encodings (0 or 1). This difference forms the basis for LogNet's superior interpretability. We highlight two key observations from Figure 6:

**Case 1 (orange bounded boxes):** For RPs 5 and 6, the DNN-DownSample model produces nearly identical and noisy latent space encodings, making it hard to tell the two classes apart or to identify which APs matter. LogNet, however, shows two clear bit differences at bit positions 12 and 46 (highlighted in pink), which separate the RP classes 5 and 6. These bits can be traced back to the APs that most influence the decision. This answers **RQ2**: LogNet helps identify which APs are most influential for distinguishing between classes, allowing us to trace decisions and apply focused fixes. In contrast, the continuous and opaque encodings from DNN-DownSample make this kind of analysis difficult, despite its popularity in indoor localization.

**Case 2 (red bounded boxes):** We now analyze RPs 59 and 60 to understand how non-ED noise affects model predictions. As discussed in Section III, real-world temporal noise is non-ED. This breaks a key assumption that underpins many DNNs, including DNN-DownSample. In DNN-DownSample, we observe small variations at latent bit positions 0, 10, 66, 70, 71, and 77 (we highlight only a subset of bits for brevity, as we observe similar trends across the latent space) between the two RPs (highlighted in green). The model appears to treat these differences as meaningful. However, these bits are not truly discriminative. If non-ED noise happens to affect these positions (0, 10, 66, 70, 71, and

77) the DNN may misclassify the input, interpreting irrelevant variations as class-defining. Worse, DNN-DownSample offers no way to verify whether these bits reflect true signal or spurious noise, due to its continuous value encoding. In contrast, LogNet exposes the correct structure of the decision boundary. It shows that class differences for RPs 59 and 60 occur at bit positions 6, 41, 42, 49, 61, and 81 (we highlight only a subset of bits for brevity, as we observe similar trends across the latent space; highlighted in blue). These critical bits consistently differ between the two classes. LogNet's rule-based logic gates ignore changes in the unimportant bits unless the noise exceeds a strict threshold, which filters out non-ED noise and focuses only on the APs that truly matter. Because each bit corresponds to a specific AP rule, the model's decisions can be traced to the APs responsible. This answers **RQ3**: LogNet improves robustness by filtering out variations from non-ED noise, and uniquely enables traceability of the most relevant APs for each class decision—features that DNNs lacks.

### D. Comparision Against Prior Works

Next, we evaluate the impact of hidden layer depth on the performance of the best-performing LogNet variant—LogNet-NOR—against the baseline DNN-DownSample model. As shown in Figure 7, models with 1 to 4 hidden layers (1H to 4H) are compared based on mean localization error averaged across all devices, RPs, and CIs (CI:0–CI:9) for both buildings. For DNNs, DNN-DownSample serves as the 1H case, while AUTLOC (2H) [21], SANGRIA (3H) [20], and DNNLOC (4H) [15] represent models from prior works with increasing hidden layers. In Figure 7, the dark black line shows mean localization error, and the whiskers represent best and worst-case performance.

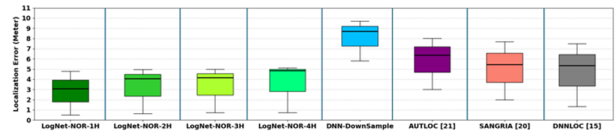

Figure 7: Impact of varying hidden layers across LogNet and DNNs.

TABLE 2: COMPARISON OF MODEL PARAMETERS AND MODEL LATENCY.

| Model | Total params | Model size (KB) | Model latency (ms) |
|---|---|---|---|
| LogNet-NOR-1H | 15374 | 60.06 | 601 |
| LogNet-NOR-2H | 7871 | 30.75 | 577 |
| LogNet-NOR-3H | 4028 | 15.74 | 567 |
| LogNet-NOR-4H | 2198 | 8.59 | 393 |
| DNN-DownSample | 56027 | 218.86 | 934 |
| AUTLOC [21] | 66286 | 228.43 | 1094 |
| SANGRIA [20] | 76546 | 299.01 | 1223 |
| DNNLOC [15] | 95292 | 372.2 | 1451 |

LogNet-NOR maintains stable mean errors across all depths, with only a slight increase as hidden layers grow. This is due to its aggressive feature compression at each layer, reducing both latent space size and total model parameters, as shown in Table 2. In contrast, DNN models improve accuracy by adding layers, but this comes at the cost of much larger model sizes, higher latency, and reduced interpretability. Specifically, LogNet-NOR-1H achieves 1.7× to 2.8× lower mean localization errors, 3.6× to 6.1× smaller model size, and 1.5× to 2.4× lower inference latency compared to all DNN variants. Even at 4 hidden layers, LogNet-NOR-4H delivers 1.1× to 1.7× lower errors, 3.4× to 43.3× smaller size, and 2.3× to 3.6× lower latency. These results demonstrate LogNet's scalability making it highly suitable for mobile deployments.

## VI. CONCLUSIONS

In this study, we introduced LogNet, a novel logic gate-based neural architecture designed to both interpret the decision-making behavior of DNNs in Wi-Fi fingerprinting-based indoor localization and enhance their long-term robustness. By analyzing temporal variation-induced noise, we demonstrated that conventional DNNs fail to capture the true non-Euclidean noise structure observed in real-world environments. LogNet addresses this gap by applying rule-based binary encoding, enabling AP-level traceability and revealing how specific noise patterns disrupt latent decision boundaries. Experimental results confirm that LogNet not only interprets internal model behavior but also delivers significant performance gains—achieving up to 1.1× to 2.8× lower localization error, 3.4× to 43.3× smaller model size, and 1.5× to 3.6× faster inference latency compared to prior DNN-based architectures over 2 years of temporal variations.